\documentclass[oneside,a4paper,onecolumn,11pt]{article}

\usepackage{fullpage}
\usepackage[left=2cm,top=2cm,bottom=2cm,right=2cm,includehead,nomarginpar,headheight=16pt]{geometry}
\usepackage[ruled, linesnumbered, vlined, commentsnumbered]{algorithm2e}
\usepackage{graphicx,subfig} 
\usepackage{amsfonts,amssymb,amsmath,amsthm,amsopn,mathrsfs,mathtools,wasysym,bbm}	
\usepackage{booktabs,diagbox,colortbl,multirow,tabularx,threeparttable,hhline}
\usepackage[listings,skins,breakable]{tcolorbox}
\usepackage{color,xcolor} 
\usepackage{fancyhdr,fancyheadings,nopageno,lastpage} 

\usepackage{enumerate}
\usepackage[shortlabels]{enumitem}
\usepackage{csquotes}
\usepackage{authblk}
\usepackage{footnote}
\usepackage{hyperref}
\usepackage{prettyref}
\usepackage{cite}
\usepackage{setspace}
\usepackage{color}
\usepackage{xcolor}  
\usepackage{geometry}
\usepackage{academicons}
\usepackage{fontawesome5}
\usepackage{pifont,ifsym,marvosym,manfnt} 
\usepackage{arydshln}
\usepackage{tikz}
\usepackage{dashrule}

\usepackage{tikz}
\usepackage{pgfplots}
\usetikzlibrary{positioning,shapes,shadows,arrows,calc}
\tikzstyle{component}=[rectangle, draw=black, rounded corners, fill=blue!40, drop shadow, text centered, anchor=north, text=white, minimum height=1cm]
\tikzstyle{arrow}=[->, thick]

\pgfplotsset{compat=1.12}
\usetikzlibrary{intersections}
\usetikzlibrary{pgfplots.statistics}
\usepgfplotslibrary{fillbetween}

\fancypagestyle{plain}{%
    \fancyfoot[C]{}
    \fancyfoot[R]{\faLeanpub\ \, \thepage\ / \pageref{LastPage}}
    \fancyfoot[L]{\faBraille\ \textsc{COLALab Report} \faSlackHash\ \footnotesize\textifsym{2024002}}
}

\hypersetup{
    colorlinks=true,
    linkcolor=ultramarine,
    filecolor=magenta,
    urlcolor=ultramarine,
    pdftitle={Overleaf Example},
    pdfpagemode=FullScreen,
}

\definecolor{red(munsell)}{rgb}{0.95, 0.0, 0.24}
\definecolor{navyblue}{RGB}{0, 0, 128}
\definecolor{myblue}{RGB}{34,31,217}
\definecolor{mycyan}{gray}{.7}
\definecolor{Gray}{gray}{0.9}
\definecolor{usccardinal}{rgb}{0.6, 0.0, 0.0}
\definecolor{ultramarine}{RGB}{0,32,96}
\definecolor{amber}{rgb}{1.0, 0.49, 0.0}

\newtheorem{remark}{Remark}

\newtheorem{definition}{Definition}

\newtcolorbox{quotebox}{colback=gray!10,boxrule=0.4pt,colframe=black,fonttitle=\bfseries,top=1pt,bottom=1pt}

\SetCommentSty{mycommfont}

\DeclareMathOperator*{\argmin}{argmin}

\newcommand{\pref}{\prettyref}
\newcommand{\our}{\texttt{EADMM}}

\newrefformat{fig}{Fig.~\ref{#1}}
\newrefformat{tab}{Table~\ref{#1}}
\newrefformat{sec}{Section~\ref{#1}}
\newrefformat{alg}{Algorithm~\ref{#1}}
\newrefformat{property}{Property~\ref{#1}}
\newrefformat{theorem}{Theorem~\ref{#1}}
\newrefformat{definition}{Definition~\ref{#1}}
\newrefformat{corollary}{Corollary~\ref{#1}}
\newrefformat{lemma}{Lemma~\ref{#1}}
\newrefformat{conj}{Conjecture~\ref{#1}}
\newrefformat{def}{Definition~\ref{#1}}
\newrefformat{eq}{equation~(\ref{#1})}
\newrefformat{app}{Appendix~\ref{#1}}
\newrefformat{remark}{Remark~\ref{#1}}

\usepackage{lscape}

\newenvironment{code-example}
{
\vspace{0.15cm}
\noindent\begin{minipage}{\linewidth}
\begin{center}
\arrayrulecolor{black}
\color{black}
\begin{tabular}{|p{0.95\linewidth}|}
\hline%
\rowcolor{pink!20}%
}
{
\\\hline
\end{tabular}
\end{center}
\end{minipage}
\vspace{-0.2cm}
}

\begin{document}

\title{\vspace{-1ex}\LARGE\textbf{Evolutionary Alternating Direction Method of Multipliers for Constrained Multi-Objective Optimization with Unknown Constraints}}

\author[1]{\normalsize Shuang Li}
\author[2]{\normalsize Ke Li}
\author[1]{\normalsize Wei Li}
\author[1]{\normalsize Ming Yang}
\affil[1]{\normalsize Control and Simulation Center, Harbin Institute of Technology, 150001, Harbin, China}
\affil[2]{\normalsize Department of Computer Science, University of Exeter, EX4 4QF, Exeter, UK}
\affil[\Faxmachine\ ]{\normalsize \texttt{frank@hit.edu.cn}}

\date{}
\maketitle

\vspace{-3ex}
{\normalsize\textbf{Abstract:}}Constrained multi-objective optimization problems (CMOPs) pervade real-world applications in science, engineering, and design. Constraint violation has been a building block in designing evolutionary multi-objective optimization algorithms for solving constrained multi-objective optimization problems. However, in certain scenarios, constraint functions might be unknown or inadequately defined, making constraint violation unattainable and potentially misleading for conventional constrained evolutionary multi-objective optimization algorithms. To address this issue, we present the first of its kind evolutionary optimization framework, inspired by the principles of the alternating direction method of multipliers that decouples objective and constraint functions. This framework tackles CMOPs with unknown constraints by reformulating the original problem into an additive form of two subproblems, each of which is allotted a dedicated evolutionary population. Notably, these two populations operate towards complementary evolutionary directions during their optimization processes. In order to minimize discrepancy, their evolutionary directions alternate, aiding the discovery of feasible solutions. Comparative experiments conducted against five state-of-the-art constrained evolutionary multi-objective optimization algorithms, on $120$ benchmark test problem instances with varying properties, as well as two real-world engineering optimization problems, demonstrate the effectiveness and superiority of our proposed framework. Its salient features include faster convergence and enhanced resilience to various Pareto front shapes.



\section{Introduction}
\label{sec:introduction}
Constrained multi-objective optimization problems (CMOPs) are ubiquitous in real-world applications. They are found in a broad spectrum of fields, including, but not limited to, tackling complex problems in computational science~\cite{ThurstonDSS03}, streamlining design and control in engineering systems~\cite{Andersson03}, and aiding in economic decision-making and strategy~\cite{PonsichJC13}. Despite their prevalence, CMOPs often present formidable challenges, primarily due to the interplay between unknown or poorly defined constraints and the requirement to balance multiple, potentially conflicting, objectives. For example, in applying reinforcement learning to autonomous driving, avoiding hazardous events like collisions is a critical constraint~\cite{GarciaF15}. The safety-critical nature of such applications makes quantifying constraint violations impractical, often leading to binary outcomes of compliance or violation. Consequently, we introduce the concept of CMOPs with unknown constraints (CMOP/UC), a topic that, to the best of our knowledge, has not been adequately addressed in the evolutionary computation community.

Evolutionary algorithms (EAs), due to their inherent population-based nature, have seen widely accepted as an effective means for multi-objective optimization. Without loss of generality, three foundational algorithms—elitist non-dominated sorting genetic algorithms (NSGA-II)\cite{DebAPM02}, multi-objective evolutionary algorithms based on decomposition (MOEA/D)\cite{ZhangL07}, and indicator-based EAs (IBEA)~\cite{ZitzlerK04}—have significantly contributed to the body of literature in the evolutionary multi-objective optimization (EMO) domain. Building on these, substantial progress has been made in developing constraint handling techniques (CHTs) for CMOPs. Initial strategies, as seen in~\cite{FonsecaF98} and~\cite{CoelloC99}, predominantly favored feasible solutions, but this approach often loses its selection pressure when no feasible solution exist—a common occurrence in scenarios with a narrow feasible region. Subsequently, many studies have focused on using constraint violation (CV) information to guide evolutionary search, incorporating CV into the Pareto dominance relation~\cite{DebAPM02,JainD14,OyamaSF07,TakahamaS12,GengZHW06,FanLCHLL16}, or developing modified environmental selection mechanisms that balance convergence, feasibility—assessed by CV—while maintaining population diversity~\cite{JimenezGSD02,RayTS01,Young05,LiDZK15,LiCFY19,TianZXZJ20,Long14,PengLG17}. A separate line of research has looked into designing specialized repair operators that adjust infeasible solutions using CV values~\cite{YuYLYC18,XuDZL20,HaradaSOK06,JiaoLSL14,SinghRS10}. However, these techniques predominantly rely on CV information, which poses limitations for CMOP/UC scenarios. In these cases, CV information might be unavailable or poorly defined, as highlighted in our recent empirical study~\cite{LiLL22}. This limitation is particularly pronounced in highly constrained feasible regions, where prevalent CHTs struggle to generate feasible solutions.

In the field of Bayesian optimization (BO)~\cite{Garnett23}, a well established method for black-box optimization, significant efforts have been made in addressing optimization problems with unknown constraints. A notable approach within this realm is the expected improvement (EI) with constraints, which views constraint satisfaction as a probabilistic classification challenge~\cite{SchonlauWJ98,GelbartSA14,JacobMZKJ14}. Beyond EI, constrained BO has seen the incorporation of various acquisition functions, such as integrated conditional EI~\cite{BernardoBBDHSW11}, expected volume reduction~\cite{Picheny14}, and predictive entropy search~\cite{LobatoHG14,Jos15}. These have shown promising results. Additionally, some algorithms employ the augmented Lagrangian function, reducing reliance on feasible solutions during the initial BO process. This approach allows the statistical model to integrate global information while the augmented Lagrangian provides localized control~\cite{GramacyGLLRWW16,PichenyRSS16,SetarehJDJ19}. Nevertheless, given these methods were originally designed to address single-objective optimization problems, even in our recent work~\cite{WangL24}, they are hardly directly applicable to deal with CMOP/UC.

Inspired by the principles of the alternating direction method of multipliers (ADMM)~\cite{BoydPCPE11}, which effectively decouples objective functions and constraints, we introduce a novel framework for tackling CMOP/UC. This framework, to our knowledge the first of its kind and hereafter referred to as evolutionary ADMM (\our), is outlined as follows.
\begin{itemize}
    \item In \our, the CMOP/UC is reformulated into two subproblems in additive form, each managed by its own evolutionary population. These populations operate in complementary evolutionary directions, a strategy empirically shown to effectively navigate highly constrained infeasible regions.

    \item A key feature of \our\ is the strategy to minimize discrepancies between the co-evolving populations by alternating their evolutionary directions. This approach fosters the discovery of feasible solutions while minimizing objective functions.

    \item The \our\ framework is universally compatible, allowing integration of any existing EMO algorithm in a plug-and-play manner. For a proof-of-concept purpose, we will apply it within three iconic algorithms, including NSGA-II, IBEA, and MOEA/D.

    \item Our rigorous evaluation of \our\ included $120$ benchmark test problems and two real-world engineering challenges. The results showcase \our's effectiveness compared to five high-performance peer algorithms.
\end{itemize}

In the rest of this paper, we will provide some preliminary knowledge pertinent to this paper along with a pragmatic review of existing works on constrained multi-objective optimization in~\pref{sec:preliminaries}. In~\pref{sec:method}, we will delineate the algorithmic implementations of our proposed \our\ framework. The experimental settings are given in~\pref{sec:settings}, and the empirical results are presented and discussed in~\pref{sec:experiments}. At the end, \pref{sec:conclusions} concludes this paper and sheds some lights on future directions.


\section{Preliminaries}
\label{sec:preliminaries}
This section starts with some basic definitions pertinent to this paper followed by a gentle tutorial of ADMM. At the end, we give a pragmatic overview of some selected up-to-date developments of EAs for CMOPs.

\subsection{Basic Definitions}
\label{sec:concepts}

The CMOP/UC considered in this paper is defined as:
\begin{equation}
    \begin{array}{ll}
        \underset{\mathbf{x}\in\Omega}{\mathrm{minimize}}\ & \mathbf{F}(\mathbf{x})=\left(f_{1}(\mathbf{x}),\ldots,f_m(\mathbf{x})\right)^\top \\
        \mathrm{subject\ to} & \mathbf{G}(\mathbf{x})=(\underbrace{0,\ldots,0}_\ell)^\top
    \end{array},
    \label{eq:cmop}
\end{equation}
where $\Omega=[x_i^L,x_i^U]^n_{i=1}\subseteq\mathbb{R}^n$ is the search space, and $\mathbf{x}=(x_1,\ldots,x_n)^\top$ is a candidate solution. $\mathbf{F}:\Omega\rightarrow\mathbb{R}^m$ consists of $m$ conflicting objective functions, and $\mathbb{R}^m$ is the objective space. $\mathbf{G}(\mathbf{x})=(g_1(\mathbf{x}),\ldots,g_\ell(\mathbf{x}))^\top:\Omega\rightarrow\mathbb{R}^\ell$ consists of $\ell\ge 1$ constraint functions. Note that this paper assumes $g_i(\mathbf{x})$ only returns a crispy response $0$ when $\mathbf{x}\in\Omega$ is feasible, otherwise it returns $1$ where $i\in\{1,\ldots,\ell\}$.

\begin{definition}
    Given two feasible solutions $\mathbf{x}^1$ and $\mathbf{x}^2$, $\mathbf{x}^1$ is said to \underline{\textit{Pareto dominate}} $\mathbf{x}^2$ (denoted as $\mathbf{x}^1\preceq\mathbf{x}^2$) if and only if $f_i(\mathbf{x}^1)\leq f_i(\mathbf{x}^2)$, $\forall i\in\{1,\ldots,m\}$ and $\exists j\in\{1,\ldots,m\}$ such that $f_
   j(\mathbf{x}^1)<f_j(\mathbf{x}^2)$.
   \label{def:ParetoDominate}
\end{definition}

\begin{definition}
    A solution $\mathbf{x}^{\ast}\in\Omega$ is \underline{\textit{Pareto-optimal}} with respect to (\ref{eq:cmop}) if $\nexists\mathbf{x}\in\Omega$ such that $\mathbf{x}\preceq\mathbf{x}^{\ast}$. 
\end{definition}

\begin{definition}
    The set of all Pareto-optimal solutions is called the \underline{\textit{Pareto-optimal set}} (PS). Accordingly, $PF=\{\mathbf{F}(\mathbf{x})|\mathbf{x}\in PS\}$ is called the \underline{\textit{Pareto-optimal front}} (PF).
\end{definition}

\begin{definition}
    The \underline{\textit{ideal objective vector}} $\mathbf{z}^\ast=(z^\ast_1,\ldots,z^\ast_m)^\top$ consists of the minimum of each objective function, i.e., $z^\ast_i=\underset{\mathbf{x}\in\Omega}{\min}\ f_i(\mathbf{x})$, $i\in\{1,\ldots,m\}$.
\end{definition}

\begin{definition}
    The \underline{\textit{nadir objective vector}} $\mathbf{z}^\mathrm{nad}=(z^\mathrm{nad}_1,\ldots,z^\mathrm{nad}_m)^\top$ consists of the worst objective function of the PF, i.e., $z^\mathrm{nad}_i=\underset{\mathbf{x}\in PS}{\max}\ f_i(\mathbf{x})$, $i\in\{1,\ldots,m\}$.
\end{definition}

\subsection{Working Mechanism of ADMM}
\label{sec:ADMMBackground}

This subsection starts with a gentle tutorial of the basic working mechanism of ADMM, followed by its extension for constrained optimization with unknown constraints. Given an equality-constrained convex optimization problem\footnote{Note that ADMM is also applicable for non-convex optimization problems~\cite{BoydPCPE11}.}:
\begin{equation}
    \begin{array}{ll}
        \underset{\mathbf{x}\in\Omega}{\mathrm{minimize}}\ & f(\mathbf{x})\\
        \mathrm{subject\ to} & A\mathbf{x}=\mathbf{b}
    \end{array},
    \label{eq:convex_problem}
\end{equation}
where $A\in\mathbb{R}^{\ell\times n}$ and $\mathbf{b}\in\mathbb{R}^\ell$. ADMM considers splitting $f(\mathbf{x})$ into the following additive form:
\begin{equation}
    \begin{array}{ll}
        \underset{\mathbf{x}\in\Omega,\ \mathbf{y}\in\Omega}{\mathrm{minimize}}\ &p(\mathbf{x})+q(\mathbf{y})\\
        \mathrm{subject\ to} & A\mathbf{x}+B\mathbf{y}=\mathbf{c}
    \end{array},
    \label{eq:admm}
\end{equation}
where $\mathbf{y}=(y_1,\ldots,y_n)^\top\in\Omega$ is an auxiliary variable, $B\in\mathbb{R}^{\ell\times n}$, and $\mathbf{c}\in\mathbb{R}^\ell$. The variable of~(\ref{eq:convex_problem}) has been split into two parts, i.e., $\mathbf{x}$ and $\mathbf{y}$, with $f(\mathbf{x})$ being separable across this splitting. In particular, $p(\mathbf{x})$ and $q(\mathbf{x})$ are assumed to be convex. To handle the constraints in~(\ref{eq:admm}), ADMM constructs the augmented Lagrangian function (ALF) as an unconstrained surrogate of~(\ref{eq:admm}):
    \begin{equation}
    \begin{aligned}
        {L_\rho}(\mathbf{x},\mathbf{y},\boldsymbol{\lambda})\buildrel\Delta\over=\ &p(\mathbf{x})+q(\mathbf{y})+{\boldsymbol{\lambda}^\top}(A\mathbf{x}+B\mathbf{y}-\mathbf{c})\\
        &+\frac{\rho}{2}\left\|{A\mathbf{x}+B\mathbf{y}-\mathbf{c}}\right\|_2^2
        \label{eq:admm_dual}
    \end{aligned},
    \end{equation}
    where $\boldsymbol{\lambda}=(\lambda_1,\ldots,\lambda_d)^\top\in\mathbb{R}^d$ is the Lagrange multiplier vector with regard to the constraints of~\pref{eq:admm}. $\left\|{A\mathbf{x}+B\mathbf{y}-\mathbf{c}}\right\|_2^2$ constitutes the penalty term where $\rho>0$ is the penalty parameter and $\|\cdot\|_2$ is the Euclidean norm. Instead of solving~(\ref{eq:admm}), ADMM works on the ALF by updating $\mathbf{x}$, $\mathbf{y}$ and $\boldsymbol{\lambda}$ in an iterative manner. During the $(k+1)$-th iteration, ADMM executes the following three steps.
\begin{equation*}
    \begin{aligned}
        \text{Step 1:}\ 
        \mathbf{x}^{k+1}&=\underset{\mathbf{x}}{\argmin}\ L_\rho\left(\mathbf{x}, \mathbf{y}^k, \boldsymbol{\lambda}^k\right) \\
        &=\underset{\mathbf{x}}{\argmin}\ p(\mathbf{x})+(\boldsymbol{\lambda}^k)^\top\left(A \mathbf{x}+B \mathbf{y}^k-\mathbf{c}\right)\\
        &\quad +\frac{\rho}{2}\left\|A \mathbf{x}+B \mathbf{y}^k-\mathbf{c}\right\|_2^2. \\
        \text{Step 2:}\ 
        \mathbf{y}^{k+1}&=\underset{\mathbf{y}}{\argmin}\ L_\rho\left(\mathbf{x}_{k+1},\mathbf{y},\boldsymbol{\lambda}^k\right) \\
        &=\underset{\mathbf{y}}{\argmin}\ q(\mathbf{y})+(\boldsymbol{\lambda}^k)^\top\left(A \mathbf{x}^{k+1}+B \mathbf{y}-\mathbf{c}\right)\\ &\quad +\frac{\rho}{2}\left\|A \mathbf{x}^{k+1}+B \mathbf{y}-\mathbf{c}\right\|_2^2. \\
        \text{Step 3:}\ 
        \boldsymbol{\lambda}^{k+1}&=\boldsymbol{\lambda}^k+\rho\left(A \mathbf{x}^{k+1}+B \mathbf{y}^{k+1}-\mathbf{c}\right).
        \label{eq:admm_iteration}
    \end{aligned}
\end{equation*}

Let us further consider a constrained optimization problem follows the problem formulation in~(\ref{eq:cmop}):
\begin{equation}
    \begin{array}{ll}
        \underset{\mathbf{x}\in\Omega}{\mathrm{minimize}}\ & f(\mathbf{x})\\
        \mathrm{subject\ to} & \mathbf{G}(\mathbf{x})=(\underbrace{0,\ldots,0}_\ell)^\top
    \end{array}.
    \label{eq:cop}
\end{equation}
According to the formulation in~(\ref{eq:admm}), we introduce an auxiliary variable $\mathbf{y}^i\in\Omega$ for $g_i(\mathbf{x})$ where $i\in\{1,\ldots,\ell\}$. Then, (\ref{eq:cop}) can be written into the following separable form:
\begin{equation}
    \begin{array}{ll}
        \underset{\mathbf{x}\in\Omega,\ \mathbf{y}^i\in\Omega}{\mathrm{minimize}}\ 
        & f(\mathbf{x})+\sum_{i=1}^\ell\left[\xi\mathbbm{1}\left(g_i\left(\mathbf{y}^i\right)==0\right)\right]\\
        \mathrm{subject\ to} & \mathbf{x}=\mathbf{y}^i,\ i\in\{1,\ldots,\ell\}
    \end{array},
    \label{eq:admm_cop}
\end{equation}
where $\xi>0$ is a scaling factor. $\mathbbm{1}(\cdot)$ is an indicator function that returns $1$ when its argument is true, otherwise it returns $0$. Following the formulation in~(\ref{eq:admm_dual}), the ALF for~(\ref{eq:admm_cop}) is constructed as:
\begin{equation}
    \begin{aligned}
        {L_\rho}(\mathbf{x},\mathbf{y}^i,\boldsymbol{\lambda}^i)\buildrel\Delta\over=\ &f(\mathbf{x})+\sum_{i=1}^\ell\big[\xi\mathbbm{1}\left(g_i\left(\mathbf{y}^i\right)==0\right)\\
        &+{\boldsymbol{\lambda}^{i\top}}(\mathbf{x}-\mathbf{y}^i)+\frac{\rho}{2}\left\|{\mathbf{x}-\mathbf{y}^i}\right\|_2^2\big]
        \label{eq:admm_cop_dual}
    \end{aligned}.
\end{equation}
Thereafter, ADMM uses the three-step procedure to solve this ALF in an iterative manner. As discussed in~\pref{sec:introduction}, ADMM was originally proposed for single-objective optimization. Its extension to CMOP/UC and its synergy with EA will be elaborated in~\pref{sec:method}.

\subsection{Literature Review of EAs for CMOPs}
\label{sec:related_works}

Since CMOP/UC have not been studied in the EMO community, we focus on the existing developments of using EAs for CMOPs whose CV information is accessible. In particular, the existing works are organized into the following five categories. Interested readers are referred to some excellent survey papers (e.g.,~\cite{FanFLLCW17, TanabeO17, SnymanH17, LiangBYQQYCT22}) for more details.

The first category mainly leverages the feasibility information during the evolutionary search process. In particular, it always grants feasible solutions a higher priority to survive to the next iteration. For example, Fonseca and Flemming~\cite{FonsecaF98} proposed a unified framework for CMOPs that prioritizes the constraint satisfaction over the optimization of objective functions. Coello Coello and Christiansen proposed a na\"ive constraint handling method that simply dumps the infeasible solutions~\cite{CoelloC99}.

The second category mainly augments the CV information to the environmental selection of EMO. In~\cite{DebAPM02}, Deb et al. proposed a constrained dominance relation to replace the vanilla Pareto dominance relation. This enables the dominance-based EMO algorithms such as NSGA-II~\cite{LiKCLZS12} and NSGA-III~\cite{JainD14} to be readily applicable to tackle CMOPs. A similar idea is applied as an alternative criterion for updating subproblems in MOEA/D variants~\cite{JanZ10,JainD14,LiKD15,ChengJOS16,LiFKZ14,LiZKLW14,WuKZLWL15,LiKZD15,LiDZK15,LiDZZ17,WuLKZZ17,WuLKZ20,WuKJLZ17,WuLKZZ19,PruvostDLL020}. Besides, the constrained dominance relation is further augmented with other heuristics to provide additional selection pressure to infeasible solutions whose CV values have a marginal difference. For example, the number of violated constraints~\cite{OyamaSF07}, $\epsilon$-constraint from the classic multi-objective optimization~\cite{TakahamaS12,MartinezC14,AsafuddoulaRS15}, stochastic ranking from the single-objective optimization~\cite{GengZHW06,YingHHLW16,LiuWW21}, and the angle between different solutions~\cite{FanLCHLL16}.

In addition to the above feasibility-driven CHTs, the third category is designed to balance the trade-off between convergence and feasibility during the evolutionary search process. One of the most popular methods along this line of research is to use the CV information to augment a penalty term to the objective function. This helps transform the constrained optimization problem into an unconstrained one. In~\cite{JimenezGSD02}, Jiménez et al. proposed a min-max formulation that drives feasible and infeasible solutions to evolve towards optimality and feasibility, respectively. In~\cite{RayTS01}, a Ray-Tai-Seow algorithm was proposed to leverage the interplay of the objective function values and the CV information to compare and rank non-dominated solutions. Based on the similar rigor, some modified ranking mechanisms (e.g.,~\cite{Young05, AngantyrAA03, WoldesenbetYT09}) were developed by taking advantage of the information from both the objective and constraint spaces. Instead of prioritizing feasible solutions, some researchers (e.g.,~\cite{LiDZK15,PengLG17,SorkhabiAK17}) proposed to exploit useful information from infeasible ones in case they can provide additional diversity to the current evolutionary population.

As a step further, the fourth category aims at simultaneously balancing convergence, diversity, and feasibility during the evolutionary search process. As a pioneer along this line, Li et al. proposed to use two co-evolving and complementary populations, denoted as two-archive EA (C-TAEA), for solving CMOPs~\cite{LiCFY19,ShanL21}. In particular, one archive, denoted as the convergence-oriented archive (CA), pushes the population towards the PF; while the other one, denoted as the diversity-oriented archive, complements the behavior of the CA and provides as much diversified information as possible. After the development of C-TAEA, the multi-population strategy has become a popular algorithmic framework for handling CMOPs (e.g.,\cite{TianZXZJ20,WangLZZG21,LiuWT21}). Besides, another idea is to transform the CMOP into an unconstrained counterpart by augmenting the constraints as additional objectives. For example,\cite{Long14} directly uses the convergence, diversity, and feasibility of obtained solutions as the auxiliary objectives to guide the evolutionary population.~\cite{PengLG17} takes the CV degree as an additional objective and assigns different weights to feasible and infeasible solutions to guide the evolutionary population to move towards the promising regions.

Instead of working on the environmental selection, the fifth category focuses on developing bespoke reproduction operators that repair infeasible solutions. For example, \cite{YuYLYC18} proposed different mutation operators to generate offspring from infeasible and feasible solutions, respectively. Likewise, Xu et al. proposed a differential evolution variant by using an infeasible-guiding mutation for offspring reproduction~\cite{XuDZL20}. In~\cite{HaradaSOK06}, a Pareto descent repair operator was proposed to explore possible feasible solutions along the gradient information around infeasible one in the infeasible region. In~\cite{JiaoLSL14}, a feasible-guided strategy was developed to guide infeasible solutions towards the feasible region along the direction starting from an infeasible solution and ending at its closest feasible solution. In~\cite{SinghRS10}, a simulated annealing was applied to accelerate the progress of movements from infeasible solutions towards feasible ones.

The last category consists of some hybrid strategies that otherwise hardly belong to any of the above categories. In~\cite{CaraffiniNP14, DattaGSD17, HernandezSDE18}, some mathematical programming methods were applied as a local search to improve the performance of EAs for solving CMOPs with equality constraints. Another idea is to split the evolutionary search process into multiple stages. For example,~\cite{FanLCLWZDG19} proposed a push and pull search framework that drives the population to jump over the infeasible region and move towards the PF without considering the constraints at the push stage. Thereafter, an improved $\epsilon$-constraint method is applied to guide the population move towards the feasible PF at the pull stage. In~\cite{TianZSZTJ21}, an adaptive two-stage EA was proposed to first drive the population towards the feasible region. Then, the population is guided to be spread along the feasible boundary in the second stage. In~\cite{SinghRS10,SinghCFD14} and~\cite{DattaR16}, surrogate-assisted EAs were proposed to handle problems with computationally expensive objective functions.

\begin{remark}
    Although the algorithms in the first category do not leverage the CV information during the evolutionary search process, the evolutionary population can easily lose selection pressure when it is filled with infeasible solutions. This is not uncommon when tackling problems with a narrow feasible region. The loss of selection pressure may lead to premature convergence and negatively affect the algorithm's ability to explore the solution space effectively. 
\end{remark}

\begin{remark}
    After the development of the constrained dominance relation~\cite{DebAPM02}, the CV information has become one of the building blocks of most, if not all, prevalent CHTs of the EMO algorithms falling into the second and third categories. As reported in our recent study~\cite{LiLL22}, it is surprising to see that the performance of this type of methods is not significantly downgraded when the CV information is not accessible. This can be attributed to the overly simplified infeasible regions that do not pose significant challenges to the existing CHTs. The robustness of these methods might also stem from the inherent adaptability of EAs and their ability to cope with various challenges.
    \label{remark:remark2}
\end{remark}

\begin{remark}
    As for the repair operators and hybrid strategies belonging to the last two categories, they are also tied with the CV information. In particular, they need to take advantage of either the latent information of CV, such as gradient, to guide the offspring reproduction towards the feasible boundary, or the feasibility information to pull infeasible solutions back to the feasible region. When dealing with unknown constraints, the reliance on CV information may limit the applicability of these strategies and hinder their performance in identifying feasible and optimal solutions.
\end{remark}


\section{Proposed Algorithm}
\label{sec:method}

\begin{figure*}[t!]
    \centering
    \includegraphics[width=0.9\textwidth]{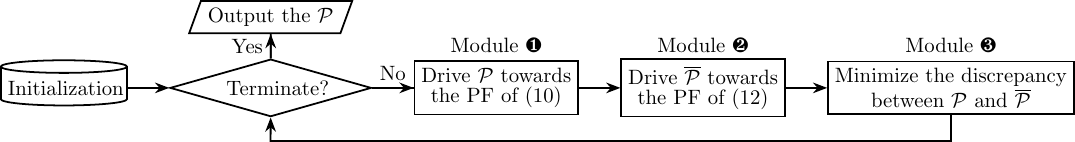}
    \caption{Flowchart of our proposed \texttt{EADMM} flowchart.}
    \label{fig:flowchart}
\end{figure*}

The flowchart of our proposed \our\ framework is outlined in~\pref{fig:flowchart}, similar to \texttt{C-TAEA}, \texttt{EADMM} maintains two co-evolving populations and consists of three algorithmic components. One (denoted as $\mathcal{P}$) is used to search for the PF of (\ref{eq:cmop}) (i.e., Module~\ding{182}); while the other one (denoted as $\mathcal{\overline{P}}$) is devoted to exploring the search space without considering any constraint (i.e., Module~\ding{183}). Furthermore, a local search is developed to minimize the discrepancy between $\mathcal{P}$ and $\mathcal{\overline{P}}$ (i.e., Module~\ding{184}). In the following paragraphs, we first reformulate the CMOP/UC into an additive form as in ADMM. Then, we delineate each algorithmic component of \texttt{EADMM} step by step.

\subsection{Problem Reformulation}
\label{sec:definition}

First of all, we rewrite (\ref{eq:cmop}) to the following form:
\begin{equation}
    \begin{array}{ll}
          {\mathrm{minimize}} & \mathbf{p}(\mathbf{x})+\mathbf{q}(\mathbf{x}),\\ 
          \mathrm{subject\ to} & \mathbf{G}(\mathbf{x})=(\underbrace{0,\ldots,0}_\ell)^\top,
    \end{array}
    \label{eq:Rewrite1}
\end{equation}
where both $\mathbf{p}(\mathbf{x})$ and $\mathbf{q}(\mathbf{x})$ are essentially $\mathbf{F}(\mathbf{x})$. In this case, (\ref{eq:Rewrite1}) is equivalent to (\ref{eq:cmop}). According to the problem formulation of ADMM in (\ref{eq:admm}), we introduce an auxiliary variable $\mathbf{y}\in\Omega$ to (\ref{eq:Rewrite1}) thus it is further rewritten as: 
\begin{equation}
   \begin{array}{ll}
      \underset{\mathbf{x}\in\Omega,\ \mathbf{y}\in\Omega}{\mathrm{minimize}} \quad & \mathbf{p}(\mathbf{x})+\mathbf{q}(\mathbf{y})\\
      \mathrm{subject\ to} & \mathbf{G}(\mathbf{x})=(\underbrace{0,\ldots,0}_\ell)^\top,\ \mathbf{x}
      =\mathbf{y}.
   \label{eq:Rewrite2}
   \end{array}
\end{equation}

\begin{remark}
    In (\ref{eq:Rewrite2}), the optimization of $\mathbf{p}(\mathbf{x})$ is equivalent to the CMOP/UC defined in (\ref{eq:cmop}) while the optimization of $\mathbf{q}(\mathbf{y})$ is the variant of (\ref{eq:cmop}) without considering the constraints $\mathbf{G}(\mathbf{x})$.
\label{remark:Px}
\end{remark}

\begin{remark}
	The constraint $\mathbf{x}=\mathbf{y}$ in (\ref{eq:Rewrite2}) is designed to leverage the information from the \textit{pseudo non-dominated solutions} without considering the constraints $\mathbf{G}(\mathbf{x})$ to guide the search for the Pareto-optimal solutions of (\ref{eq:cmop}). In this case, we can expect the pseudo non-dominated solutions may be infeasible.
\end{remark}

\subsection{Implementation of Module~\ding{182}}
\label{sec:EADMM}

According to~\pref{remark:Px}, the purpose of Module~\ding{182} shown in~\pref{fig:flowchart} is to guide $\mathcal{P}$ to approximate the PF of the following optimization problem:
\begin{equation}
   \begin{array}{ll}
      \underset{\mathbf{x}\in\Omega}{\mathrm{minimize}} \quad & \mathbf{p}(\mathbf{x})\\ 
      \mathrm{subject\ to} & \mathbf{G}(\mathbf{x})=(\underbrace{0,\ldots,0}_\ell)^\top, 
   \end{array}
   \label{eq:EADMM_Step2}
\end{equation}
where $\mathbf{p}(\mathbf{x})$ is $\mathbf{F}(\mathbf{x})$ which makes~(\ref{eq:EADMM_Step2}) be equivalent to~(\ref{eq:cmop}). To handle unknown constraints, we propose to modify the conventional CHTs, mainly based on the CV information as reviewed in~\pref{sec:related_works}, to adapt to the Pareto-, indicator-, and decomposition-based EMO frameworks. As examples, here we choose three iconic algorithms \texttt{NSGA-II}~\cite{DebAPM02}, \texttt{IBEA}~\cite{ZitzlerK04}, and \texttt{MOEA/D}~\cite{ZhangL07} as the representative algorithm for each framework, respectively.

\subsubsection{Modification on NSGA-II} Here we  only need to modify the dominance relation. Specifically, a solution $\mathbf{x}^1$ is said to \underline{constraint-dominate} $\mathbf{x}^2$, if: 1) $\mathbf{x}^1$ is feasible while $\mathbf{x}^2$ is not; 2) both of them are infeasible while the number of constraints violated by $\mathbf{x}^1$ is less than that of $\mathbf{x}^2$; 3) both of them are infeasible and satisfy the same number of constraints while $\mathbf{x}^1\preceq\mathbf{x}^2$ without considering constraints; or 4) both of them are feasible while $\mathbf{x}^1\preceq\mathbf{x}^2$.

\subsubsection{Modification on IBEA} For \texttt{IBEA}, we only need to consider feasible solutions in the environmental selection. Note that if none of the offspring solutions are feasible, all parents survive to the next iteration.

\subsubsection{Modification on MOEA/D} To adapt \texttt{MOEA/D} to handle CMOP/UC, we only need to modify the solution update mechanism. Specifically, given an offspring solution $\mathbf{x}^c$, it can replace the selected parent $\tilde{\mathbf{x}}$ if one of the following four conditions is met: 1) $\mathbf{x}^c$ is feasible while $\tilde{\mathbf{x}}$ is not; 2) both of them are infeasible while the number of constraints violated by $\mathbf{x}^c$ is less than that of $\tilde{\mathbf{x}}$; 3) both of them are infeasible and meet the same number of constraints while $ g^{\operatorname{tch}}\left(\mathbf{x}^c \mid \mathbf{w}, \tilde{\mathbf{z}}^\ast\right)\leq g^{\operatorname{tch}}\left(\tilde{\mathbf{x}}\mid\mathbf{w},\tilde{\mathbf{z}}^\ast\right)$; or  4) both of them are feasible and $ g^{\operatorname{tch}}\left(\mathbf{x}^c \mid \mathbf{w}, \tilde{\mathbf{z}}^\ast\right)\leq g^{\operatorname{tch}}\left(\tilde{\mathbf{x}}\mid\mathbf{w},\tilde{\mathbf{z}}^\ast\right)$. In particular, $g^{\operatorname{tch}}\left(\cdot\mid\mathbf{w},\tilde{\mathbf{z}}^\ast\right)$ is the widely used Tchebycheff function~\cite{LiZKLW13}:
\begin{equation}
    g^{\operatorname{tch}}\left(\cdot \mid \mathbf{w}, \tilde{\mathbf{z}}^*\right) = \max _{1 \leq i \leq m} \frac{\left|f_i(\cdot)-\tilde{z}^*\right|}{w_i},
    \label{eq:Tchebyche}
\end{equation}
where $\mathbf{w}=\left(w_1,\ldots,w_m\right)^{\top}$ and $\sum_{i=1}^m w_i=1$ is the weight vector of the subproblem associated with $\tilde{\mathbf{x}}$, and $\tilde{\mathbf{z}}^*$ is the estimated ideal point according to the current evolutionary population, where $\tilde{z}_i^*=\underset{{\mathbf{x}\in\mathcal{P}\cup\left\{\mathbf{x}^c\right\}}}{\min}f_i(\mathbf{x})$, $i\in\{1,\ldots,m\}$.

\begin{remark}
    Note that the above modifications are so minor that the corresponding innate algorithmic frameworks are kept intact. In addition, they are applicable to any other variants of \texttt{NSGA-II}, \texttt{IBEA}, and \texttt{MOEA/D} in a plug-in manner.
\end{remark}

\subsection{Implementation of Module~\ding{183}}

The optimization problem considered in Module~\ding{183} is:
\begin{equation}
   \underset{\mathbf{y} \in \Omega}{\operatorname{minimize}}  \quad \mathbf{q}(\mathbf{y}),
   \label{eq:EADMM_Step3}
\end{equation}
where $\mathbf{q}(\mathbf{y})$ is essentially $\mathbf{F}(\mathbf{x})$ as defined in~\pref{sec:definition}. To solve this problem, any off-the-shelf EMO algorithm can be applied without any modification. Here we choose the vanilla \texttt{NSGA-II}, \texttt{IBEA}, and \texttt{MOEA/D} to keep the algorithmic framework of Module~\ding{183} consistent with Module~\ding{182}.
\begin{remark}
   Module~\ding{183} is designed to enable $\overline{\mathcal{P}}$ to explore the search space as much as possible without being restricted by the feasibility. This helps provide a necessary diversity that promotes the interplay between $\mathcal{P}$ and $\overline{\mathcal{P}}$ as in Module~\ding{184}.
\end{remark}

\subsection{Implementation of Module~\ding{184}}

The purpose of Module~\ding{184} is to minimize the discrepancy between $\mathcal{P}$ and $\overline{\mathcal{P}}$ thus to satisfy the constraint $\mathbf{x}=\mathbf{y}$ defined in~(\ref{eq:Rewrite2}). To this end, we need to take the offspring populations generated in both Module~\ding{182} and Module~\ding{183}, denoted as $\mathcal{Q}$ and $\overline{\mathcal{Q}}$ respectively, into consideration. It consists of the following five consecutive steps.
\begin{enumerate}[Step 1:]
   \item Use $\overline{\mathcal{Q}}$ to update $\mathcal{P}$ according to the environmental selection mechanism of the backbone algorithm used in Module~\ding{182}.
   \item  Use $\mathcal{Q}$ to update $\overline{\mathcal{P}}$ according to the environmental selection mechanism of the backbone algorithm used in Module~\ding{183}.
   \item Create a temporary archive $\hat{\mathcal{S}}=\left\{\hat{\mathbf{x}}^i\right\}_{i=1}^{|\hat{\mathcal{S}}|}$ where $\hat{\mathbf{x}}^i \in \mathcal{Q}$. In addition, $\exists\mathbf{x}\in\mathcal{P}\wedge\exists \overline{\mathbf{x}}\in\overline{\mathcal{P}}$ such that $\hat{\mathbf{x}}^i\preceq\mathbf{x}\wedge\hat{\mathbf{x}}^i\preceq\overline{\mathbf{x}}$ according to~\pref{def:ParetoDominate}. The constraints $\mathbf{G}(\mathbf{x})=(\underbrace{0,\ldots,0}_\ell)^\top$ are not considered. 
   \item For each solution $\hat{\mathbf{x}}^i \in \hat{\mathcal{S}}$, solve the following optimization problem:
   \begin{equation}
       \check{\mathbf{x}}^i=\underset{\mathbf{x} \in \Omega}{\operatorname{argmin}} \sum_{j=1}^{\ell} \mathbbm{1}\left(g_j(\mathbf{x})==0\right)+\rho\left\|\mathbf{x}-\hat{\mathbf{x}}^i\right\|_2^2
       \label{eq:local_search}
   \end{equation}
   where $i\in\{1,\ldots,|\hat{\mathcal{S}}|\}$ and $\rho=\frac{\gamma}{|\mathcal{P}|}\times\ell$, $\gamma$ is the number of feasible solutions in $\mathcal{P}$. 
   \item Use $\check{\mathbf{x}}^i$ where $i\in\{1,\ldots,|\hat{\mathcal{S}}|\}$ to update $\mathcal{P}$ and $\overline{\mathcal{P}}$ according to the environmental selection mechanism of the corresponding backbone algorithm in Modules~\ding{182} and~\ding{183}, respectively.
\end{enumerate}

\begin{figure}[t!]
    \centering
    \includegraphics[width=0.8\columnwidth]{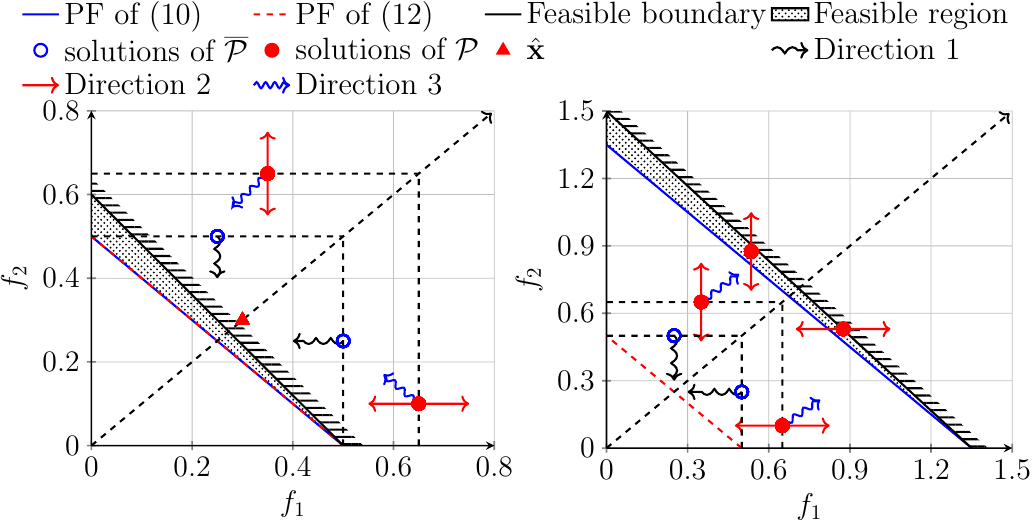}
    \caption{Illustrations of the working mechanism of EADMM/MOEAD, where $\dashrightarrow$ is the weight vector, \hdashrule[0.5ex]{0.5cm}{0.5pt}{2pt} is the contour line of the Tchebycheff function.}
    \label{fig:module3_example}
\end{figure}
\begin{remark}
    To elucidate the underpinnings of Module~\ding{184}, let us first consider an example shown in~\pref{fig:module3_example}(a) where the PF of~\eqref{eq:EADMM_Step2} overlaps with that of~\eqref{eq:EADMM_Step3}. Here, $\overline{\mathcal{P}}$ will be impelled towards the PF of~\eqref{eq:EADMM_Step3} under the guidance of Module~\ding{183} (denoted as $\rightsquigarrow$). Meanwhile, $\mathcal{P}$ prioritizes the search for feasible solutions, steered by Module~\ding{182}. When the feasible region is very narrow, $\mathcal{P}$ may flounder in the infeasible region (denoted as \textcolor{red}{$\longleftrightarrow$}). This can be attributed to the limited selection pressure that can mislead the search towards less promising areas. To mitigate this, Steps 1, 2, and 5 of Module~\ding{184} will drive $\mathcal{P}$ to move towards $\overline{\mathcal{P}}$ (denoted as \textcolor{blue}{$\rightsquigarrow$}). Since the PF of~\eqref{eq:EADMM_Step2} overlaps with the PF of~\eqref{eq:EADMM_Step3}, it signifies that solutions adhering to the condition $\mathbf{x}=\mathbf{y}$ are within the solution space of~\eqref{eq:EADMM_Step3}, enabling $\mathcal{P}$ to locate these solutions under the navigation of $\overline{\mathcal{P}}$. Moreover, when an offspring $\hat{\mathbf{x}}$ (denoted as \textcolor{red}{$\blacktriangle$}) derived from $\mathcal{P}$ dominates both $\mathcal{P}$ and $\overline{\mathcal{P}}$, Step 4 comes into play, executing a local search centered around $\hat{\mathbf{x}}$.
    \label{remark:EADMM_Illustration1}
\end{remark}

\begin{remark}
    On the other hand, considering the example shown in the~\pref{fig:module3_example}(b), wherein the PF of~\eqref{eq:EADMM_Step2} does not overlap with~\eqref{eq:EADMM_Step3}, the solutions of $\mathcal{P}$ located in the feasible region will continue to explore the feasible region guided by Module~\ding{182} (denoted as \textcolor{red}{$\longleftrightarrow$}). Whereas the solutions in $\mathcal{P}$ that follow $\overline{\mathcal{P}}$ and move into the infeasible region will be pulled back to the feasible region by the operations performed in Module~\ding{182} (denoted as \textcolor{blue}{$\rightsquigarrow$}). In such cases, the function of Steps 1, 2, and 5 in Module~\ding{184} is twofold: to track the evolutionary trajectory of $\overline{\mathcal{P}}$ and to consider the outcomes of the local search conducted in Step 4. When superior solutions meeting the constraint $\mathbf{x}=\mathbf{y}$ are identified, they prompt $\mathcal{P}$ to shift towards the area of the search space where these advanced solutions are located.
    \label{remark:EADMM_Illustration2}
\end{remark}

\begin{remark}
    In theory, any off-the-shelf optimization algorithm can be applied to solve~\eqref{eq:local_search}. In this paper, we choose the genetic algorithm toolbox of the MATLAB 2021a for a proof-of-concept purpose.
\end{remark}

\begin{remark}
    Step 4 serves as a local search around $\hat{\mathbf{x}}^i$ where $i\in\{1,\ldots,|\hat{\mathcal{S}}|\}$. This helps search for feasible solutions with regard to~(\ref{eq:EADMM_Step2}) and satisfy the constraint $\mathbf{x}=\mathbf{y}$ defined in~(\ref{eq:Rewrite2}).
\end{remark}

\begin{remark}
    Due to the consideration of the number of feasible solutions in $\mathcal{P}$, it makes the local search conducted in the Step 4 be adaptive. Specifically, a smaller $\gamma$ tends to search for feasible solutions in a larger region close to $\hat{\mathbf{x}}^i$; whereas a larger $\gamma$ leads to the search of feasible solutions in a smaller region close to $\hat{\mathbf{x}}^i$ where $i \in\{1,\ldots,|\hat{\mathcal{S}}|\}$.
\end{remark}

\subsection{Time Complexity Analysis}
\label{sec:complexity}

The time complexity of our proposed algorithm, which consists of three algorithmic components, is analyzed separately for each module. For both Modules~\ding{182} and~\ding{183}, their computational complexity is consistent with their corresponding backbone algorithms. Specifically, the complexity is $\mathcal{O}(mN^2)$ for \texttt{NSGA-II}, $\mathcal{O}(N^2)$ for \texttt{IBEA}, and $\mathcal{O}(mTN)$ for \texttt{MOEA/D}, where $N$ is the size of $\mathcal{P}$ and $\overline{\mathcal{P}}$, and $T$ is the number of weight vectors in the neighborhood of each weight vector of \texttt{MOEA/D}. As for Module~\ding{184}, Steps 1, 2, and 5 employ the same environmental selection mechanism as Modules~\ding{182} and~\ding{183}, resulting in a time complexity that aligns with their corresponding backbone algorithms. In Step 4, since the optimization of~\eqref{eq:local_search} is performed $|\hat{\mathcal{S}}|$ times, the computational complexity amounts to $\mathcal{O}\left(|\hat{\mathcal{S}}|\tilde{N}^2\right)$, where $\tilde{N}$ is the population size used in the genetic algorithm. In summary, the overall complexity of our proposed \our\ framework is $\max\left\{N^2,\mathcal{O}\left(|\hat{\mathcal{S}}|\tilde{N}^2\right)\right\}$.


\section{Experimental Setup}
\label{sec:settings}
This section introduces the settings of our empirical study including the benchmark test problems, the peer algorithms, the performance metrics, and the statistical tests.

\subsection{Benchmark Test Problems}
\label{sec:benchmark}

Our empirical study considers both synthetic test problems and real-world engineering challenges. We will briefly introduce their characteristics in the following paragraphs.

\subsubsection{Synthetic Test Problems}
\label{sec:synthetic}
\begin{figure*}[t!]
    \centering
    \includegraphics[width=1\textwidth]{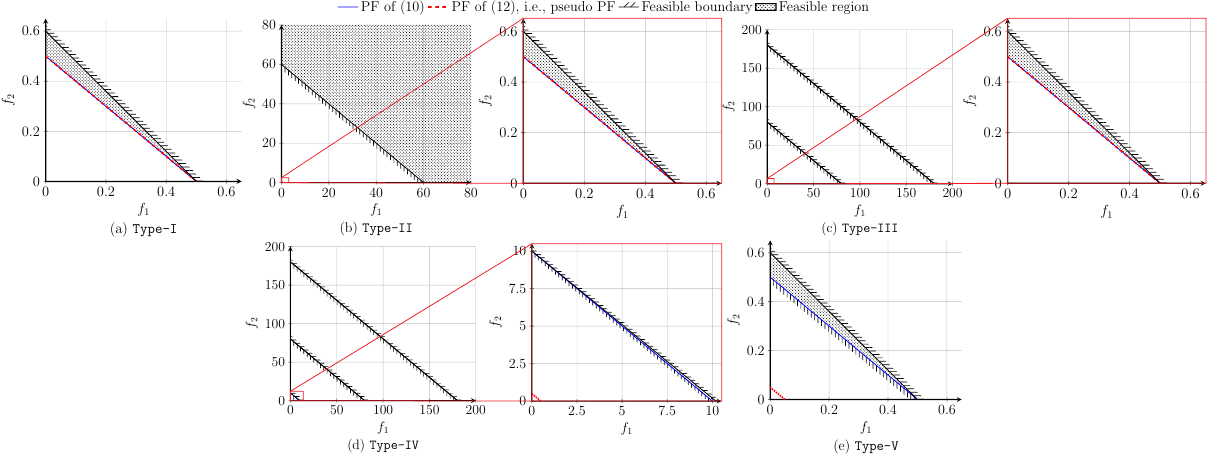}
    \caption{Illustrations for the PFs and the feasible regions of the synthetic test problems built upon C1-DTLZ1. Subfigures (a) to (e) represents the cases ranging from \texttt{Type-I} to \texttt{Type-V}.}
    \label{fig:uc_examples}
\end{figure*}

As highlighted in~\pref{remark:remark2}, existing test problems are insufficient for adequately benchmarking the existing CHTs in EMO for CMOP/UC scenarios. To address this gap, we have developed a new set of synthetic test problems, drawing inspiration from the C-DTLZ benchmark suite~\cite{JainD14}. These new problems exhibit the following five types of characteristics:
\begin{itemize}
    \item\underline{\texttt{Type-I}}: As the illustrative example shown in~\pref{fig:uc_examples}(a), the feasible region of this type of problem is a narrow area adjacent to the PF, making most of the search space infeasible. Constrained EMO algorithms relying on the CV information may struggle to navigate towards the PF through this predominantly infeasible space.

    \item\underline{\texttt{Type-II}}: An extension of \texttt{Type-I}, this category includes additional feasible regions distant from the PF, as shown in~\pref{fig:uc_examples}(b). The large gap between these disparate feasible areas often diminishes the selection pressure necessary for evolutionary populations to traverse the regions using binary CV signals.

    \item\underline{\texttt{Type-III}}: Building upon \texttt{Type-I}, this type intersperses the infeasible space with two narrowly spaced feasible regions, as depicted in~\pref{fig:uc_examples}(c). Without access to CV information, evolutionary populations may become impeded by these narrow feasible areas.

    \item\underline{\texttt{Type-IV}}: Inspired by C3-DTLZ problems, the PF here is overshadowed by a \lq pseudo\rq\ PF when constraints are disregarded, as in~\pref{fig:uc_examples}(d). This presents additional complexities that can misguide an evolutionary population towards the infeasible region above the PF.

    \item\underline{\texttt{Type-V}}: Sharing the same PF and feasible regions as \texttt{Type-I}, this category introduces the complexity of a \lq pseudo\rq\ PF similar to \texttt{Type-IV}, as shown in~\pref{fig:uc_examples}(e), elevating the challenge.
\end{itemize}

\begin{remark}
   Without loss of generality, \pref{fig:uc_examples} only uses C1-DTLZ1 as the baseline to illustrate the above five types of characteristics. Further, considering the stochastic nature of real-world problems, the constraints are with noises in the proposed synthetic test problems. The mathematical definitions and the corresponding visual illustrations of all test problem instances can be found in Section A of Appendix A of the supplemental document.
\end{remark}

\subsubsection{Real-world Problems}
\label{sec:real_world}

In addition to the synthetic test problems, we also examine two real-world optimization problems derived from different domains.
\begin{itemize}
    \item We selected the lunar lander task from Gymnasium\footnote{\url{https://gymnasium.farama.org/}} as our first test case. This task involves applying reinforcement learning (RL) to control the rocket's landing trajectories. To optimize performance, we consider fine-tuning $12$ hyperparameters associated with the RL algorithm. Our objectives are to maximize total rewards while minimizing landing duration. The task includes two unknown constraints: maintaining the lander's awakeness and avoiding crashes.

    \item The second real-world problem we address concerns water distribution systems (WDS), an indispensable and highly costly component of public infrastructure. The planning and management of WDS typically involve balancing multiple conflicting objectives, such as operational cost, system resilience, and profitability. Our experiments focus on the Pescara network (PES)~\cite{BragallDLLT08}, a well-known benchmark in WDS optimization, which comprises $99$ pipes. The optimization goal is to adjust the diameters of these pipes to improve overall system performance. Our objectives are to minimize operational costs and maximize system resilience. Additionally, we consider two unknown constraints: preventing backflow and avoiding pipe blockage.
\end{itemize}
The detailed mathematical formulation and settings of these two engineering optimization problems can be found in Section B of Appendix A of the supplemental document.

\subsection{Peer Algorithms and Parameter Settings}
\label{sec:peers_parameters}

For a proof-of-concept purpose, we use \texttt{NSGA-II}, \texttt{IBEA}, and \texttt{MOEA/D} as the backbone algorithms under our proposed \texttt{EADMM} framework. The corresponding \texttt{EADMM} instances are denoted as \texttt{EADMM}/\texttt{NSGA-II}, \texttt{EADMM}/\texttt{IBEA}, and \texttt{EADMM}/\texttt{MOEA/D}, respectively. Five high-performance EMO algorithms for CMOPs, including \texttt{C-TAEA}~\cite{LiCFY19}, \texttt{PPS}~\cite{FanLCLWZDG19}, \texttt{MOEA/D-DAE}~\cite{ZhuZL20}, \texttt{ToP}~\cite{LW19}, \texttt{CMOCSO}~\cite{MingGLWG22}, are chosen as the peer algorithms. As discussed in~\pref{sec:related_works}, these peer algorithms require accessing to the CV information which is yet available in CMOP/UC. To address this issue, we follow the practice in~\cite{LiLL22} to replace the CV with a crispy value, i.e., the CV value of a feasible solution is $0$; otherwise, it is $1$ if the solution is infeasible.

The parameter settings are listed as follows.
\begin{itemize}
    \item\underline{Number of function evaluations (FEs)}: The maximum number of FEs for different problems are listed in Table I and Table II of Appendix B of the supplemental document. All synthetic test problems are scalable to any number of objectives while we consider $m\in\{2,3,5,10\}$ in our experiments.

    \item\underline{Reproduction operators}: The parameters associated with the simulated binary crossover and polynomial mutation are set as $p_c=1.0$, $\eta_c=20$, $p_m=\frac{1}{n}$, $\eta_m=20$. As for those use differential evolution~\cite{StornP97} for offspring reproduction, we set $CR=F=0.5$.

    \item\underline{Algorithm specific parameters}: The parameter settings of peer algorithms considered in our experiments are set identical to their original papers. Note that our proposed \our\ framework does not introduce any additional parameters, except the innate ones of the backbone algorithms. All parameter settings are detailed in Table III of Appendix B of the supplemental document.

    \item\underline{Number of repeated runs}: Each algorithm is independently run on each test problem for $31$ times with different random seeds.
\end{itemize}

\subsection{Performance Metrics and Statistical Tests}
\label{sec:metrics}

We employ inverted generational distance (IGD)~\cite{BosmanT03}, IGD$^+$~\cite{IshibuchiMTN15}, and hypervolume (HV)~\cite{ZitzlerT99} in performance assessment. The reference point used in the HV evaluation is constantly set as $(\underbrace{1.1, \ldots, 1.1}_m)^\top$. Note that all these three performance metrics can assess both convergence and diversity. The smaller the IGD and IGD$^+$, or the larger the HV, the better result is achieved by the corresponding algorithm.

In view of the stochastic nature of EAs, we use the following three statistical tests to conduct a statistical interpretation of the significance of the comparison results.
\begin{itemize}
	\item\underline{Wilcoxon signed-rank test}~\cite{Frank92}: This is a non-parametric statistical test that makes no assumption about the underlying distribution of the data. It has been recommended in many empirical studies in the EA community~\cite{DerracGMH11}. The significance level is set to $p=0.05$ in our experiments.

	\item\underline{$A_{12}$ effect size}~\cite{VarghaD00}: To ensure the resulted differences are not generated from a trivial effect, we apply $A_{12}$ as the effect size measure to evaluate the probability that one algorithm is better than another. Specifically, given a pair of peer algorithms, $A_{12}=0.5$ means they are \textit{equal}. $A_{12}<0.5$ denotes that one is worse for more than 50\% of the times. $0.36\leq A_{12}<0.44$ indicates a \textit{small} effect size while $0.29\leq A_{12}<0.36$ and $A_{12}< 0.29$ mean a \textit{medium} and a \textit{large} effect size, respectively. 

	\item\underline{Scott-Knott test}: Instead of merely comparing the raw metric values, we apply the Scott-Knott test to rank the performance of different peer algorithms over $31$ runs on each test problem. In a nutshell, the Scott-Knott test uses a statistical test and effect size to divide the performance of peer algorithms into several clusters. The performance of peer algorithms within the same cluster is statistically equivalent. The clustering process terminates until no split can be made. Finally, each cluster can be assigned a rank according to the mean metric values achieved by the peer algorithms within the cluster. The smaller the rank is, the better performance of the algorithm achieves.
\end{itemize}


\section{Empirical Studies}
\label{sec:experiments}

In this section, we present and analyze the empirical results of our experiments from three aspects.

\begin{figure*}[t!]
    \centering
    \includegraphics[width=.9\textwidth]{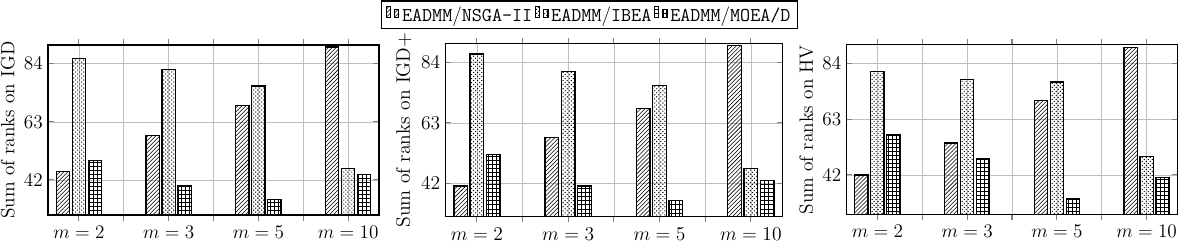}
    \caption{Total Scott-Knott test ranks achieved by each of the three algorithm instances of our proposed framework (the smaller the rank is, the better performance achieved).}
    \label{fig:Scott_EADMM}
\end{figure*}

\begin{figure*}[t!]
    \centering
    \includegraphics[width=.9\textwidth]{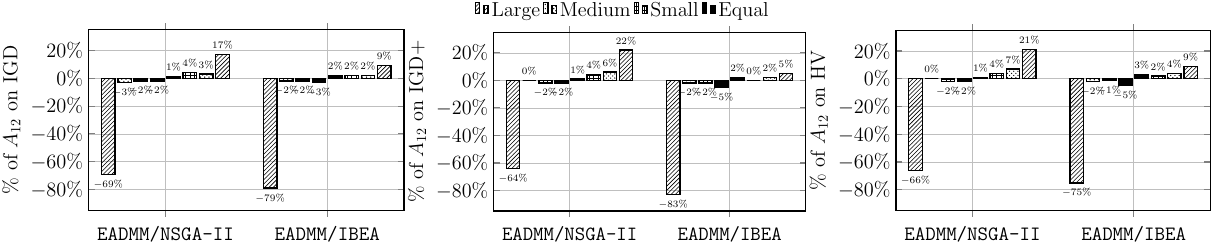}
    \caption{Percentage of the large, medium, small, and equal $A_{12}$ effect size, respectively, when comparing \texttt{EADMM/MOEA/D} against \texttt{EADMM/NSGA-II} and \texttt{EADMM/IBEA}.}
    \label{fig:Scott_EADMM}
\end{figure*}

\begin{figure*}[t!]
    \centering
    \includegraphics[width=\textwidth]{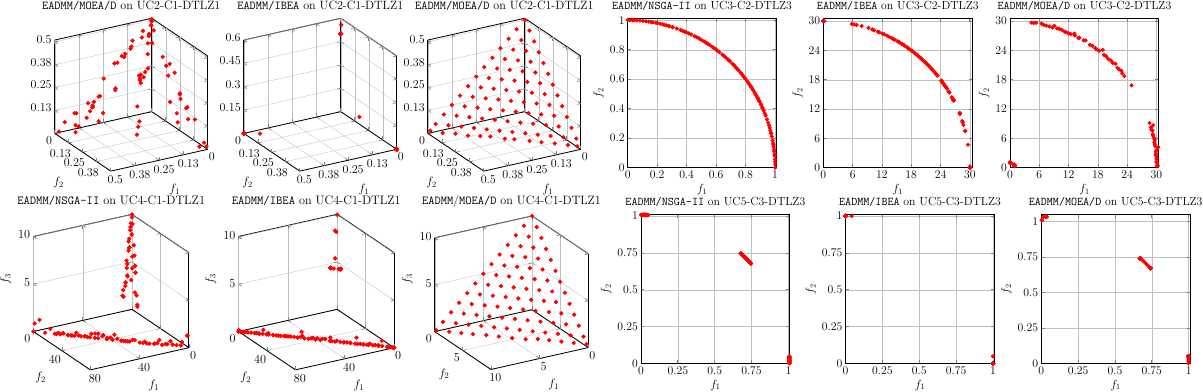}
    \caption{Scatter plots of the final solutions obtained by each of the three algorithm instances of our proposed framework on four synthetic test problems with the median IGD values.}
    \label{fig:DTLZ_EADMM}
\end{figure*}

\begin{figure}[t!]
    \centering
    \includegraphics[width=0.40\linewidth]{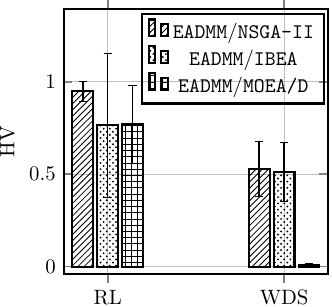}
    \caption{Bar charts with error bars of HV values obtained by each of the three algorithm instances of our proposed framework on two real-world problems.}
    \label{fig:RW_HV}
\end{figure}

\begin{figure*}[t!]
    \centering
    \includegraphics[width=\textwidth]{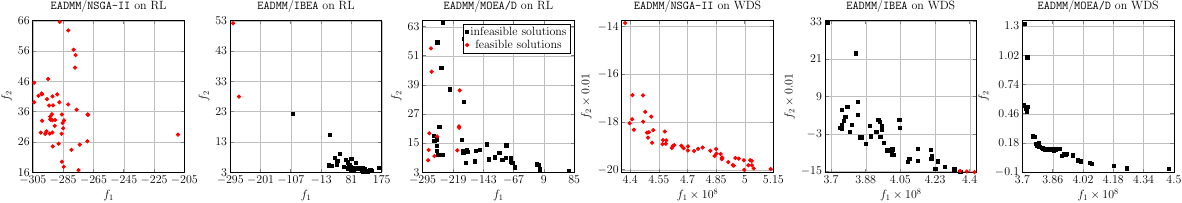}
    \caption{Scatter plots of the final solutions (with the median HV values) obtained by \texttt{EADMM/NSGA-II}, \texttt{EADMM/IBEA}, and \texttt{EADMM/MOEA/D} on two real-world problems.}
    \label{fig:RW_EADMM}
\end{figure*}

\subsection{Performance Comparison of Three \our\ Instances}
\label{sec:ProposedInstances}

As introduced in~\pref{sec:peers_parameters}, we developed three algorithm instances based on the \our\ framework. The primary goal of our initial experiment is to assess the comparative performance of these instances. We present the statistical comparison results of IGD, IGD$^+$, and HV, based on the Wilcoxon signed-rank test, in Tables IV to XVI in Appendix C of the supplemental document. From these results, it is interesting to observe that no single algorithm consistently excels across all benchmark test problems.

To better understand the performance rankings among these three instances, we employ the Scott-Knott test. This test categorizes them into different groups based on IGD, IGD$^+$, and HV values for each synthetic test problem. We opted not to list all ranking results ($30\times 4\times 3=360$ in total) to avoid clutter. Instead, we summarize these results in bar charts in~\pref{fig:Scott_EADMM}, providing an overview of their comparative performance. Notably, \texttt{EADMM}/\texttt{MOEA/D} emerges as the top-performing instance in $80\%$ of comparisons. \texttt{EADMM}/\texttt{IBEA} and \texttt{EADMM}/\texttt{NSGA-II} also show strong performances in specific scenarios, although the latter's effectiveness diminishes with an increasing number of objectives.

To facilitate visual comparisons, we select some plots of the final solutions obtained by each of the three algorithm instances of our proposed framework on four synthetic test problems in~\pref{fig:DTLZ_EADMM}\footnote{\label{foot: figures}Full results of visual comparisons can be found in Fig.6 to Fig.17 of Section B of Appendix C of the supplementary document.}. It is clear to see that the solutions found by \texttt{EADMM}/\texttt{MOEA/D} exhibit the best convergence and diversity on the $3$-objective UC2-C1-DTLZ1 and UC4-C1-DTLZ1 problems. However, on $2$-objective UC3-C2-DTLZ3 and UC5-C3-DTLZ3 problems, the convergence and diversity of solutions found by \texttt{EADMM}/\texttt{MOEA/D} are worse than that found by \texttt{EADMM/NSGA-II}. This means that, on the one hand, the proposed \texttt{EADMM} framework effectively guides the populations through the infeasible regions and avoids them being misled by the pseudo-PF. On the other hand, the performance of algorithms instances of \texttt{EADMM} is significantly affected by the backbone algorithms.

Due to the absence of ground truth PF for the real-world engineering problems, we base our performance comparisons on the HV metric. The bar charts with error bars in~\pref{fig:RW_HV} show that \texttt{EADMM}/\texttt{NSGA-II} is the most competitive. We also plot the final solutions, both feasible and infeasible, in~\pref{fig:RW_EADMM}. For the lunar lander task, \texttt{EADMM}/\texttt{NSGA-II} excels in guiding the population toward feasible regions, while more than half of the final solutions obtained by \texttt{EADMM}/\texttt{IBEA} and \texttt{EADMM }/\texttt{MOEA/D} are located in the infeasible region. As for the WDS planning and management task, all solutions obtained by \texttt{EADMM}/\texttt{NSGA-II} are feasible. Although \texttt{EADMM}/\texttt{IBEA} finds some feasible solutions, most of its population are still located in the infeasible region. In contrast, \texttt{EADMM}/\texttt{MOEA/D} fails to find any feasible solution. This can be attributed to the disparate scales of the two objective functions, where $f_1$ is on the order of $10^9$ larger than $f_2$. Such an imbalance skews the evolutionary direction toward $f_1$, thereby sacrificing the population diversity and inadvertently steering the search into infeasible regions.

\subsection{Performance Comparisons with Other Five State-of-the-art Peer Algorithms}
\label{sec:PeerAlgorithms}

\begin{figure}[t!]
    \centering
    \includegraphics[width=.49\textwidth]{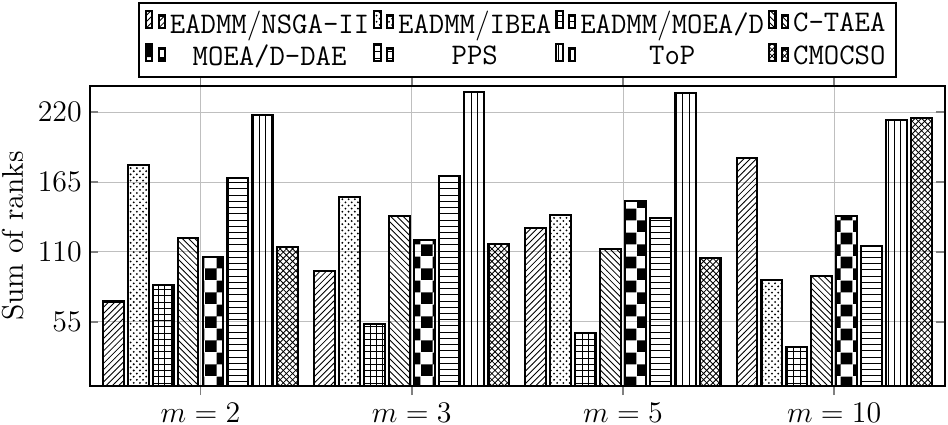}
    \caption{Total Scott-Knott test ranks achieved by each of the three proposed algorithms and other five state-of-the-art peer algorithms (the smaller the rank is, the better performance achieved).}
    \label{fig:Scott_Peers}
\end{figure}

\begin{figure*}[t!]
    \centering
    \includegraphics[width=\textwidth]{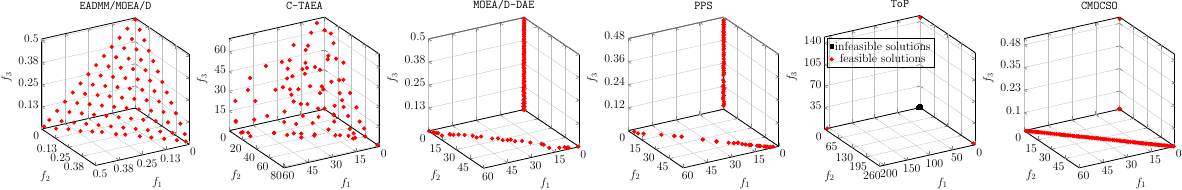}
    \caption{Final solutions found by different algorithms with the median IGD values on three objective UC2-C2-DTLZ1 test problems.}
    \label{fig:UC2-C1-DTLZ1_m3}
\end{figure*}
\begin{figure*}[t!]
    \centering
    \includegraphics[width=\textwidth]{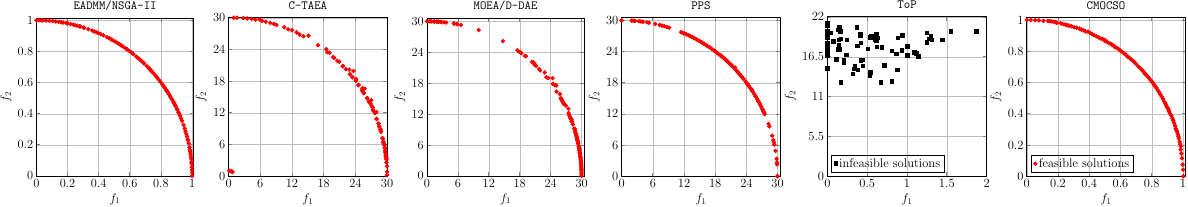}
    \caption{Final solutions found by different algorithms with the median IGD values on two objective UC3-C2-DTLZ3 test problems.}
    \label{fig:UC3-C2-DTLZ3_m2}
\end{figure*}
\begin{figure*}[t!]
    \centering
    \includegraphics[width=\textwidth]{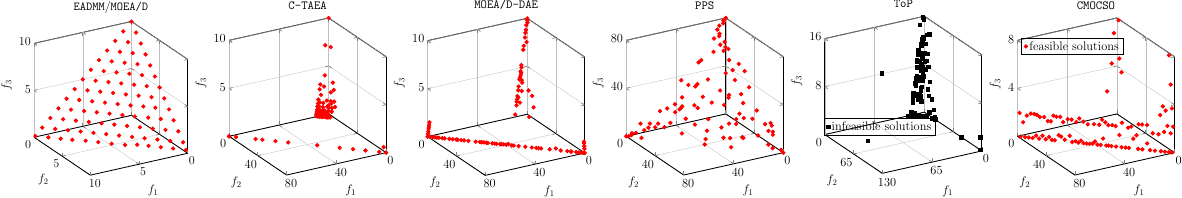}
    \caption{Final solutions found by different algorithms with the median IGD values on three objective UC4-C1-DTLZ1 test problems.}
    \label{fig:UC4-C1-DTLZ1_m3}
\end{figure*}
\begin{figure*}[t!]
    \centering
    \includegraphics[width=\textwidth]{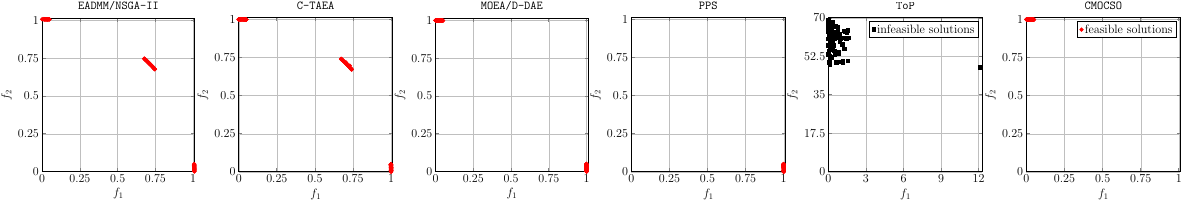}
    \caption{Final solutions found by different algorithms with the median IGD values on two objective UC5-C3-DTLZ3 test problems.}
    \label{fig:UC5-C3-DTLZ3_m2}
\end{figure*}

In this subsection, we compare the performance of all our three \our\ instances against the five state-of-the-art peer algorithms introduced in~\pref{sec:peers_parameters}. Similar to the practice conducted in~\pref{sec:ProposedInstances}, we first employ the Wilcoxon signed-rank test to compare the IGD, IGD$^+$, and HV results of each of our three algorithm instances with regard to the other five state-of-the-art peer algorithms. The statistical results are given in Tables IV to XVI of Appendix C of our supplemental document. It is interesting to note that the comparison results align with our observation in~\pref{sec:ProposedInstances}. In a nut shell, we find that \texttt{EADMM}/\texttt{MOEA/D} is the best algorithm on the synthetic test problems while it obtains significantly better metric values in $80\%$ comparisons. As for the two real-world engineering problems, \texttt{EADMM}/\texttt{NSGA-II} is the most competitive one.

As done in~\pref{sec:ProposedInstances}, we apply the Scott-Knott test to sort the performance of all algorithms on the synthetic test problem instance across different number of objectives. To facilitate a better interpretation of these massive comparison results, we summarize the test results obtained across all test problem instances for each algorithm and show them as the bar charts in~\pref{fig:Scott_Peers}. From this result, we find that two of our proposed \our\ instances secure the top two positions in almost all comparisons. This validates the superiority of the proposed \our\ framework. Specifically, like the observations in~\pref{sec:ProposedInstances}, \texttt{EADMM}/\texttt{NSGA-II} is the best algorithm when $m=2$ while \texttt{EADMM}/\texttt{MOEA/D} wins in other cases. Note that although \texttt{EADMM}/\texttt{IBEA} is relatively less competitive than its two siblings, it still is ranked in the second and third place on $m=10$ and $m=5$ test problem instances respectively. 

To have a better visual interpretation of the superiority achieved by our algorithm instances, let us look into the population distribution of the final solutions against the other five peer algorithms in~\pref{fig:UC2-C1-DTLZ1_m3} to~\pref{fig:UC5-C3-DTLZ3_m2}. Due to space limitations, only selected examples are included here, with complete results available in the supplemental document\footnote{See footnote~\pref{foot: figures}}. These plots reveal that the solutions from \texttt{EADMM}/\texttt{MOEA/D} and \texttt{EADMM}/\texttt{NSGA-II} demonstrate notable superiority over the other five peer algorithms. We will now briefly analyze the potential reasons behind this performance.
\begin{itemize}
   \item From the second columns of~\pref{fig:UC2-C1-DTLZ1_m3} to~\pref{fig:UC5-C3-DTLZ3_m2}, it is evident that \texttt{C-TAEA} demonstrates superior convergence in UC5-C3-DTLZ3 compared to the other three problems. This superior performance can be attributed to the UC5-C3-DTLZ3's feasible region being adjacent to the PF. Here, the convergence-oriented archive of \texttt{C-TAEA} is effectively steered towards the feasible region, aided by its diversity-oriented archive. In contrast, the other three problems feature multiple feasible regions not adjacent to the PF. Without CV guidance, the convergence-oriented archive's solutions are still directed towards the feasible regions. However, the update mechanism in the CA tends to enhance solution diversity within each feasible region. As a result, the convergence in these problems is not as pronounced as in UC5-C3-DTLZ3.
   
	\item The data in the third columns of~\pref{fig:UC2-C1-DTLZ1_m3} to~\pref{fig:UC5-C3-DTLZ3_m2} show that \texttt{MOEA/D-DAE} struggles with converging populations to the feasible regions on the PF for the first three problems. It only manages to locate solutions in some local feasible regions of the UC5-C3-DTLZ3. This issue arises because \texttt{MOEA/D-DAE} depends on the rate of change in CV to assess the evolutionary state of populations. It utilizes an $\varepsilon$-based CHT to direct populations towards feasible areas. However, in the absence of CV information, the algorithm's detect-and-escape strategy finds it challenging to accurately discern the evolutionary state. Consequently, the $\varepsilon$-based CHT inadvertently leads populations to linger in feasible regions that are not on the PF.	

   \item Examining the fourth columns of~\pref{fig:UC2-C1-DTLZ1_m3} to~\pref{fig:UC5-C3-DTLZ3_m2} we notice that the performance of \texttt{PPS} mirrors that of \texttt{MOEA/D-DAE}. This similarity is largely because \texttt{PPS} also utilizes an $\varepsilon$-based CHT that depends on CV information. Specifically, for the first three problems, \texttt{PPS} prematurely transitions to the pull stage, which happens before the populations reach the feasible regions on the PF. Lacking CV information, the CHT mistakenly pulls the infeasible solutions back to a non-PF feasible region. In the case of UC5-C3-DTLZ3, while \texttt{PPS} identifies solutions in some local feasible regions during the push stage, the transition to the pull stage proves ineffective. Without CV guidance, the algorithm struggles to explore other feasible regions.  

   \item As illustrated in the fifth columns of~\pref{fig:UC2-C1-DTLZ1_m3} to~\pref{fig:UC5-C3-DTLZ3_m2}, \texttt{ToP} fails to find feasible solutions for the last three problems. This limitation stems from the algorithm's reliance on CV information during the single-objective optimization stage. In the absence of CV data, the populations are erroneously directed towards infeasible regions. Consequently, when \texttt{ToP} transitions to the multi-objective optimization stage, the populations continue gravitating towards areas with better objective function fitness values, albeit infeasible. This misdirection makes it challenging for the populations to locate any feasible solutions.  
   
   \item As we can see from the sixth columns of~\pref{fig:UC2-C1-DTLZ1_m3} to~\pref{fig:UC5-C3-DTLZ3_m2}, although \texttt{CMOCSO} outperforms the other four peer algorithms, it still does not successfully find solutions in all the local feasible regions of UC5-C3-DTLZ3. Moreover, it struggles to converge all solutions to the feasible regions on the PF in both UC2-C1-DTLZ1 and UC4-C1-DTLZ1. This limitation is mainly due to the algorithm's reliance on the $\varepsilon$-based CHT. When deprived of CV information, the CHT may lead to the selection of misleading winner particles. Consequently, \texttt{CMOCSO}'s cooperative swarm optimizer faces difficulties in maintaining both the diversity and convergence of the particles.
\end{itemize}

\begin{figure}[t!]
    \centering
    \includegraphics[width=0.6\linewidth]{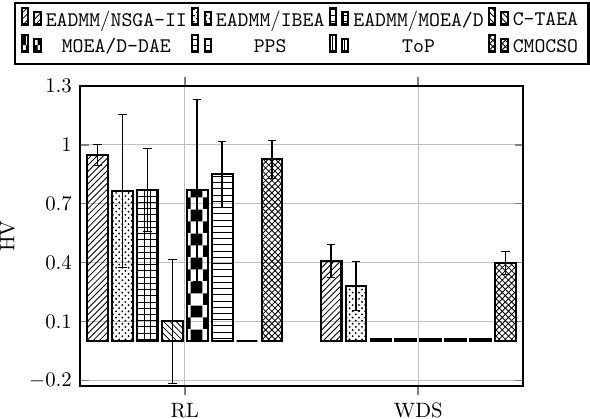}
    \caption{Bar charts with error bars of HV values obtained by each of the three algorithm instances of our proposed framework and other five state-of-the-art algorithms.}
    \label{fig:RWHV_Peers}
\end{figure}

\begin{figure*}[t!]
    \centering
    \includegraphics[width=\textwidth]{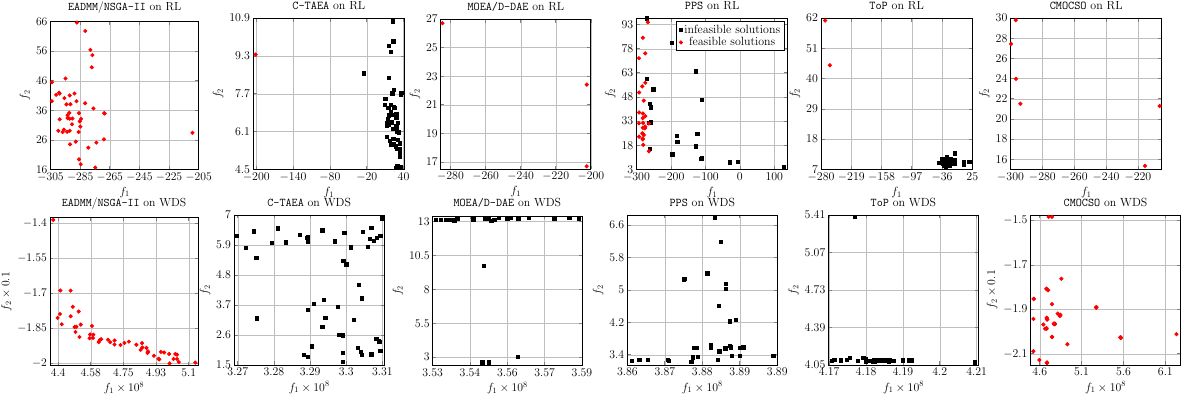}
    \caption{Final solutions found by different algorithms with the median HV values on real-world problems.}
    \label{fig:RW_Peers}
\end{figure*}

As we did in~\pref{sec:ProposedInstances}, we compare the HV values of our three \our\ instances against five state-of-the-art peer algorithms on two real-world engineering problems. The statistical results are presented as bar charts with error bars in~\pref{fig:RWHV_Peers}. Additionally, the final solutions corresponding to the median HV values are illustrated in~\pref{fig:RW_Peers}. From the bar charts in~\pref{fig:RWHV_Peers}, it is evident that \texttt{EADMM}/\texttt{NSGA-II} achieves the best HV values in both the lunar lander and the WDS planning and management tasks. This performance is further supported by the scatter plots in~\pref{fig:RW_Peers}, where \texttt{EADMM}/\texttt{NSGA-II} clearly outperforms other algorithms. Specifically, for the lunar lander task, all solutions by \texttt{EADMM}/\texttt{NSGA-II} are feasible. In the WDS task, while both \texttt{EADMM}/\texttt{NSGA-II} and \texttt{CMOCSO} successfully guide the population towards the feasible region, the prior one demonstrates superior convergence and diversity. In contrast, algorithms such as \texttt{C-TAEA}, \texttt{MOEA/D-DAE}, \texttt{PPS}, and \texttt{ToP} fail to find any feasible solutions. We will briefly analyze the potential reasons as follows.
\begin{itemize}
    \item As MOEA/D variants, \texttt{C-TAEA}, \texttt{MOEA/D-DAE}, and \texttt{PPS} struggle with significantly disparate objective functions, as discussed in~\pref{sec:ProposedInstances}. The scatter plots in~\pref{fig:RW_Peers} show that these algorithms tend to optimize $f_1$ at the expense of solution diversity. Consequently, their populations gravitate towards regions where $f_1$ is more optimal but infeasible.

    \item \texttt{ToP} is a two-stage evolutionary algorithm that initially emphasizes single-objective optimization, primarily focusing on the objective with higher fitness improvement. In the WDS planning and management task, this approach results in a biased selection pressure towards the first objective. Such bias leads to reduced population diversity and drives the population away from feasible regions. Therefore, when \texttt{ToP} transitions to the multi-objective optimization stage, it encounters difficulties in locating feasible solutions due to the diminished diversity.
\end{itemize}

\subsection{Ablation study with regard to the EADMM}
\begin{figure}[t!]
    \centering
    \includegraphics[width=0.6\linewidth]{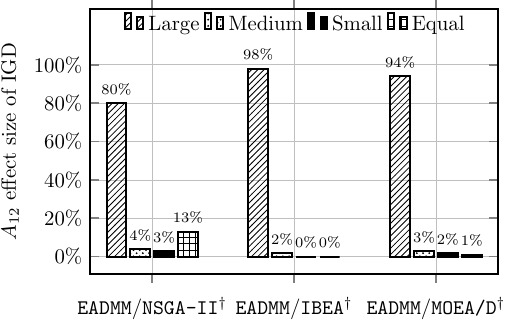}
    \caption{Percentage of the large, medium, small, and equal A12 effect size of IGD.}
    \label{fig:Ablation_A12}
\end{figure}

\begin{figure*}[t!]
    \centering
    \includegraphics[width=\textwidth]{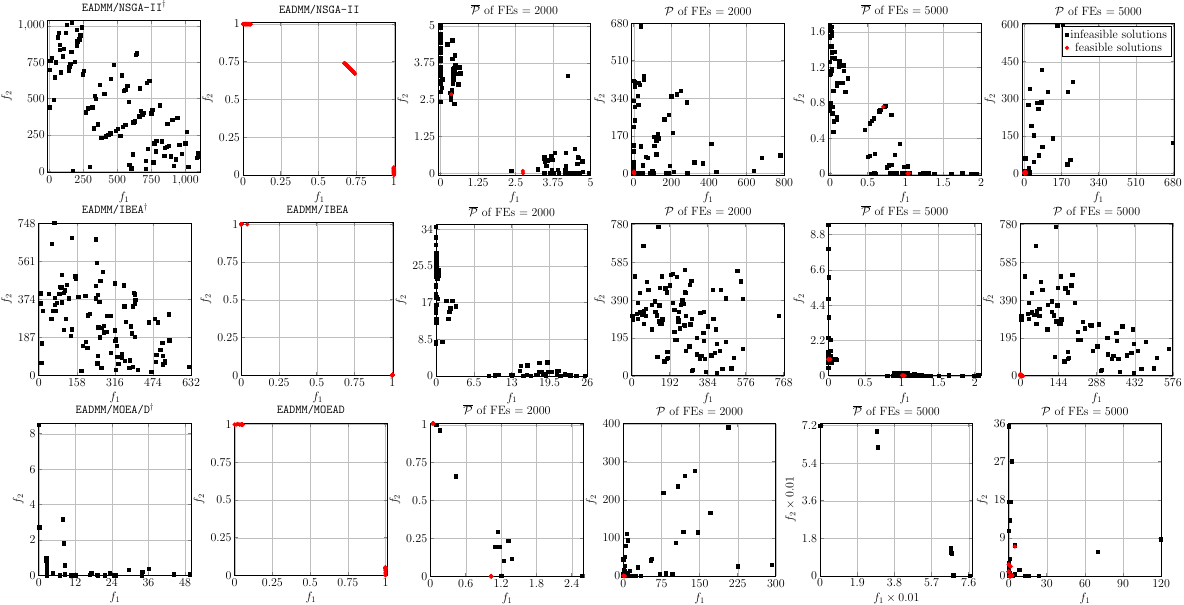}
    \caption{The first column plots are the final solutions found by \texttt{EADMM}/\texttt{NSGA-II}$^\dagger$, \texttt{EADMM}/\texttt{IBEA} $^\dagger$ and \texttt{EADMM}/\texttt{MOEA/D} $^\dagger$ with the median IGD values on UC4-C3-DTLZ2 respectively. The last five column plots are the solutions found by the corresponding algorithm instances under our proposed \texttt{EADMM} framework, the $\overline{\mathcal{P}}$ and $\mathcal{P}$ obtained by these instances using different FEs respectively.}
    \label{fig:UC4-C3-DTLZ2}
\end{figure*}

The empirical study presented in~\pref{sec:PeerAlgorithms} demonstrates the superior performance of our proposed \texttt{EADMM} framework compared to five state-of-the-art EMO algorithms. As illustrated in~\pref{fig:flowchart}, Modules \ding{183} and \ding{184} are the key distinguishing elements of our framework. We plan to assess the effectiveness of these modules through an ablation study. This study will involve comparing the performance of the complete \texttt{EADMM} framework against a version that exclusively uses Module \ding{182}, denoted by the $^\dagger$ symbol.

Based on the Wilcoxon signed-rank test, the statistical comparison of IGD, IGD$^+$, and HV values, presented in Tables XVII to XXVIII of Appendix C of the supplementary document, alongside the $A_{12}$ effect size depicted in~\pref{fig:Ablation_A12}, indicates a consistent performance degradation when Modules \ding{183} and \ding{184} are ablated. As an illustrative example, \pref{fig:UC4-C3-DTLZ2} demonstrates this point. The ablated versions, \texttt{EADMM}/\texttt{NSGA-II}$^\dagger$, \texttt{EADMM}/\texttt{IBEA}$^\dagger$, and \texttt{EADMM}/\texttt{MOEA/D}$^\dagger$, all fail to find feasible solutions on UC4-C3-DTLZ2. In contrast, the full versions of \texttt{EADMM}/\texttt{NSGA-II}, \texttt{EADMM}/\texttt{IBEA}, and \texttt{EADMM}/\texttt{MOEA/D} successfully drive their populations to converge into the feasible regions. The reasons for this phenomenon are explained as follows.
\begin{itemize}
    \item As we can see from the last four columns of plots in~\pref{fig:UC4-C3-DTLZ2}, it is noticeable that the solutions in $\overline{\mathcal{P}}$ exhibit better objective function fitness compared to $\mathcal{P}$. Additionally, $\mathcal{P}$ at $FEs=5000$ shows improved feasibility and objective function fitness over the final solutions of the algorithms marked with $\dagger$, as seen in the first column of plots. This observation suggests that $\overline{\mathcal{P}}$ effectively guides $\mathcal{P}$ towards the PF, in line with the discussion in~\pref{remark:EADMM_Illustration1} and as illustrated in the left plot of~\pref{fig:module3_example}.

    \item In the fourth column of plots in~\pref{fig:UC4-C3-DTLZ2}, some solutions of $\overline{\mathcal{P}}$ are in the infeasible region, showing better objective function fitness than those in feasible areas. However, the final solutions of $\mathcal{P}$, depicted in the second column of plots, are all within feasible regions. This indicates that solutions in $\mathcal{P}$, initially following $\overline{\mathcal{P}}$ into the infeasible region, are eventually redirected back to the feasible areas. This behavior is described in~\pref{remark:EADMM_Illustration2} and shown in the right plot of~\pref{fig:module3_example}.
\end{itemize}

Furthermore, in~\pref{fig:UC4-C3-DTLZ2}, we observe that the solutions of different algorithms have shown varying degrees of convergence and diversity. The potential reasons for these phenomena are discussed as follows.
\begin{itemize}
    \item Focusing on the first column of plots in~\pref{fig:UC4-C3-DTLZ2}, we see that the solutions by \texttt{EADMM/MOEA/D}$^\dagger$ show the best convergence in the objective space. This is mainly because Module \ding{182} in \texttt{EADMM/MOEA/D} prioritizes infeasible solutions with better objective function fitness, provided they meet the same number of constraints. In contrast, Module \ding{182} in \texttt{EADMM/IBEA} treats all infeasible solutions equally, while \texttt{EADMM/NSGA-II} prioritizes infeasible solutions based on the number of met constraints. The convergence advantage of \texttt{EADMM/MOEA/D} is also evident in the $\overline{\mathcal{P}}$ and ${\mathcal{P}}$ in the third row of plots, enhancing its performance in synthetic test problems with many objectives.

    \item In the first row of plots in~\pref{fig:UC4-C3-DTLZ2}, the $\overline{\mathcal{P}}$ and ${\mathcal{P}}$ of \texttt{EADMM/NSGA-II} demonstrate greater diversity compared to \texttt{EADMM/IBEA} and \texttt{EADMM/MOEA/D}. Consequently, \texttt{EADMM/NSGA-II} is the only algorithm among those in the second column of plots to successfully locate feasible solutions in all feasible regions on the PF. This can be attributed to its crowding distance assignment mechanism, which prevents clustering of solutions in the objective space. Such diversity grants \texttt{EADMM/NSGA-II} exceptional performance in real-world problems and synthetic test problems in the $2$-objective cases.
\end{itemize}


\section{Conclusions and Future Directions}
\label{sec:conclusions}
CMOP/UC are important in various real-world optimization scenarios, particularly in safety-critical situations. Despite their significance, CMOP/UC has been significantly underexplored in the EMO community. In this paper, we bridge this gap by synergizing traditional ADMM with evolutionary meta-heuristics to propose the \our\ framework. This approach is simple yet effective in tackling CMOP/UC challenges. Our comprehensive experiments, conducted on a range of synthetic test problems and real-world engineering optimization scenarios, validate the robustness and effectiveness of the \our\ framework. Further, our ablation study also demonstrates the vital roles of the ADMM problem formulation (i.e., Modules \ding{183} and \ding{184} in \our), reinforcing the framework's performance capabilities. As a pioneering work in the field, we had an eureka moment that the implications of CMOP/UC and the \our\ framework extend far beyond the scope of this paper, there are several open questions remained.
\begin{itemize}
    \item \textit{Theoretical analysis}: The theoretical properties of ADMM have been rigorously studied in the class optimization literature. It will be interesting to leverage the mathematical underpinnings of ADMM to gain a deeper theoretical understanding of \our\ in terms of its optimization behavior and convergence properties.

    \item \textit{Scalability studies}: Further research into the scalability of the \our\ framework, particularly in handling problems with a large number of variables~\cite{WilliamsLM21}, would be beneficial. This could include the development of parallelized or distributed versions of the framework to leverage computational resources more effectively. It is also interesting to extend the static environments to non-stationary environments~\cite{ChenLY18,FanLT20,ChenL21a,LiCY23}. Constraints can be represented as another format of preferences, to which our group has made a series of contributions towards solutions of interest in a priori~\cite{CaoKWL14,LiDAY17,LiCMY18}, a posterior~\cite{ZouJYZZL19,NieGL20,ChenL21b,LiNGY22}, and interactive~\cite{Li19,LiCSY19,LiLY23} manners, as well as performance benchmarking~\cite{LiDY18,LaiLL21,TanabeL23}. In particular, our previous study has empirically demonstrated the effectiveness of leveraging user preferences in the search of solutions of interest~\cite{LiLDMY20}.

    \item \textit{Real-world applications}: Expanding the application of the \our\ framework to more diverse and complex real-world problems, such as large-scale environmental modeling or advanced engineering design, would be an exciting avenue. This requires a multi-disciplinary collaboration to ensure our research to be aligned with the practical needs and challenges of different domains, such as natural language processing~\cite{YangL23}, neural architecture search~\cite{ChenL23,LyuYWHL23,LyuHYCYLWH23,LyuLHWYL23}, robustness of neural networks~\cite{ZhouLM22a,ZhouLM22b,WilliamsLM23a,WilliamsLM23b,WilliamsLM22,WilliamsL23c}, software engineering~\cite{LiXT19,LiXCWT20,LiuLC20,LiXCT20,LiYV23}, smart grid management~\cite{XuLAZ21,XuLA21,XuLA22}, communication networks~\cite{BillingsleyLMMG19,BillingsleyLMMG20,BillingsleyMLMG20,BillingsleyLMMG21}, machine learning~\cite{CaoKWL12,LiWKC13,LiK14,CaoKWLLK15,WangYLK21}, and visualization~\cite{GaoNL19}.
\end{itemize}

\section*{Acknowledgment}
K. Li was supported in part by the UKRI Future Leaders Fellowship under Grant MR/S017062/1 and MR/X011135/1; in part by NSFC under Grant 62376056 and 62076056; in part by the Royal Society under Grant IES/R2/212077; in part by the EPSRC under Grant 2404317; in part by the Kan Tong Po Fellowship (KTP\textbackslash R1\textbackslash 231017); and in part by the Amazon Research Award and Alan Turing Fellowship. S. Li, W. Li, and M. Yang were supported by the National Natural Science Foundation of China under Grant 62273119.

\bibliographystyle{IEEEtran}
\bibliography{IEEEabrv,your_bib}

\newpage


\end{document}